\documentclass[conference]{IEEEtran}
\IEEEoverridecommandlockouts
\newcommand\copyrighttext{%
	\footnotesize \textcopyright2018 IEEE:
	Personal use of this material is permitted. Permission from IEEE must be obtained for all other uses, in
	any current or future media, including reprinting/republishing this material for advertising or promotional
	purposes, creating new collective works, for resale or redistribution to servers or lists, or reuse of any
	copyrighted component of this work in other works.}
\newcommand\copyrightnotice{%
	\begin{tikzpicture}[remember picture,overlay]
		\node[anchor=south,yshift=10pt] at (current page.south) {\fbox{\parbox{\dimexpr\textwidth-\fboxsep-\fboxrule\relax}{\copyrighttext}}};
	\end{tikzpicture}%
}

\usepackage{cite}
\usepackage{amsmath,amssymb,amsfonts}
\usepackage{algorithmic}
\usepackage{graphicx}
\usepackage{textcomp}
\usepackage{xcolor}
\usepackage{orcidlink}
\usepackage{subfig}
\usepackage[numbers]{natbib}
\begin{document}

\title{Thermal-LiDAR Fusion for Robust Tunnel Localization in GNSS-Denied and Low-Visibility Conditions
\thanks{This work was supported by the projects ARCHIMEDES (Grant Agreement
No. 101112295) and Cynergy4MIE (Grant Agreement Nr. 101140226), which
are Co-funded by the European Union. The projects are supported by the
Chips Joint Undertaking and its members including top-up funding by the
program “Digitale Technologien” of the Austrian Federal Ministry for Innovation, Mobility and Infrastructure (BMIMI). The publication was written at Virtual Vehicle Research GmbH in Graz and partially funded within the COMET K2 Competence Centers for Excellent Technologies from the Austrian Federal Ministry for Innovation, Mobility and Infrastructure (BMIMI), Austrian Federal Ministry for Economy, Energy and Tourism (BMWET), the Province of Styria (Dept. 12) and the Styrian Business Promotion Agency (SFG). The Austrian Research Promotion Agency (FFG) has been authorised for the programme management.
Views and opinions expressed
are, however, those of the author(s) only and do not necessarily reflect those
of the European Union Key Digital Technologies Joint Undertaking. Neither
the European Union nor the granting authority can be held responsible for
them.}
}

\author{\IEEEauthorblockN{Lukas Schichler \orcidlink{0000-0002-9180-9600}}
\IEEEauthorblockA{\textit{Virtual Vehicle Research GmbH} \\
Graz, Austria \\
lukas.schichler@v2c2.at}
\and
\IEEEauthorblockN{Karin Festl\orcidlink{0000-0003-1773-2104}}
\IEEEauthorblockA{\textit{Virtual Vehicle Research GmbH} \\
Graz, Austria \\}
\and
\IEEEauthorblockN{Selim Solmaz\orcidlink{0000-0003-0686-1306}}
\IEEEauthorblockA{\textit{Virtual Vehicle Research GmbH} \\
Graz, Austria \\}
\and
\IEEEauthorblockN{Daniel Watzenig\orcidlink{0000-0002-5341-9708}}
\IEEEauthorblockA{\textit{Virtual Vehicle Research GmbH} \\
Graz, Austria \\}
}

\maketitle

\copyrightnotice
\begin{abstract}
Despite significant progress in autonomous navigation, a critical gap remains in ensuring reliable localization in hazardous environments such as tunnels, urban disaster zones, and underground structures. Tunnels present a uniquely difficult scenario: they are not only prone to GNSS signal loss, but also provide little features for visual localization due to their repetitive, textureless walls and poor lighting. These conditions degrade conventional vision-based and LiDAR-based systems, which rely on distinguishable environmental features. To address this, we propose a novel sensor fusion framework that integrates a thermal camera with a LiDAR to enable robust localization in tunnels and other perceptually degraded environments. 
The thermal camera provides resilience in low-light or smoke conditions, while the LiDAR delivers precise depth perception and structural awareness. By combining these sensors, our framework ensures continuous and accurate localization across diverse and dynamic environments.
We use an Extended Kalman Filter (EKF) to fuse multi-sensor inputs while accommodating varying sampling rates and sensor outages. The framework leverages visual odometry and SLAM (Simultaneous Localization and Mapping) techniques to process the sensor data, enabling robust motion estimation and mapping even in GNSS-denied environments. This fusion of sensor modalities not only enhances system resilience but also provides a scalable solution for cyber-physical systems in connected and autonomous vehicles (CAVs).
To validate the framework, we conduct tests in a tunnel environment, simulating sensor degradation and visibility challenges. The results demonstrate that our method sustains accurate localization where standard approaches deteriorate due to the tunnel’s featureless geometry. The framework’s versatility makes it a promising solution for autonomous vehicles, inspection robots, and other cyber-physical systems operating in constrained, perceptually poor environments.
\end{abstract}


\section{Introduction}

Autonomous navigation has made remarkable strides in recent years, yet reliable localization in hazardous or perceptually degraded environments remains a critical challenge. GNSS-denied areas, smoke, and complex structures demand resilient multi-sensor solutions. We present a robust localization system combining LiDAR and thermal camera data through an Extended Kalman Filter (EKF). This loosely-coupled fusion approach accommodates asynchronous sensor data while maintaining accurate pose estimation during complete sensor outages.

While traditional vehicle odometry relies on wheel encoders, IMUs, and GNSS, vision and LiDAR-based methods are gaining prominence. Key developments include:
\begin{itemize}
    \item LiDAR odometry: LOAM  (Lidar Odometry and Mapping) \cite{loam} matches geometric features between scans
    \item Camera odometry: DSO (Direct Sparse Odometry) \cite{dso}, ORB-SLAM \cite{orbslam}, and SVO (Semi-Direct Visual Odometry) \cite{svo}
\end{itemize}
Multi-sensor fusion expands operational capabilities by combining complementary strengths:
GNSS provides coarse global positioning, IMU (inertial measurement unit) provides accurate motion estimates, LiDAR enables precise local mapping and Thermal cameras maintain functionality in smoke or fog. There are several solutions for the coupling of LiDAR and IMU, for example, LVIO-SAM (Lidar Visual Inertial Odometry via Smoothing and Mapping)~\cite{lviosam} integrates an inertial measurement unit (IMU) with LiDAR odometry. The IMU data are used to deskew the LiDAR data, while the LiDAR data are used to correct the IMU bias. 
Other techniques, such as LVI-SLAM \cite{lvisam}, operate with separate visual inertial and LiDAR inertial subsystems that function independently during sensor failure or jointly when sufficient features are available. 
For legged robots operating on uneven terrain, IMU-based motion estimation becomes particularly challenging. The measurements are complicated by the superposition of the robot’s actual motion with various disturbances. Foot-ground impacts, leg dynamics, and terrain-induced oscillations introduce high-frequency noise, while drift effects accumulate over time. 

For reliable localization, this work focuses specifically on the fusion of LiDAR and thermal camera, as these sensors offer particularly robust performance in challenging environments.
By introducing thermal imaging alongside LiDAR measurements, our approach addresses the challenge of smoke-filled areas, featureless tunnels and low-light conditions. By employing visual methods that leverage diverse sensor data and conceptually distinct approaches (e.g., we use both photometric and feature-based methods), we maximize the robustness of the total system against influences that may severely impact any single localization unit.
The multi-sensor localization system is designed for hazardous and dynamic environments where individual sensors may intermittently fail.
In \cite{Schichler2025}, we introduced the concept of fusing thermal camera, LiDAR sensor and a GNSS system. The implemented solution showed satisfactory results in an open environment. In this work, we extend the approach to localization in a tunnel environment. As there is no GNSS signal and the tunnel provides only little features, the demands on the localization system increase. We extend the solution accordingly. To validate the effectiveness, we conduct simulations, navigating a ground vehicle through a replica of the tunnel system in Zentrum am Berg\cite{zab}. During the test, we simulate sensor failures by selectively disabling individual sensors and demonstrate the system's ability to maintain localization despite these disruptions.

In Sections \ref{sec:lidar} and~\ref{sec:thermal}, the two odometry methods for the LiDAR sensor and thermal camera are described. In Section~\ref{sec:ekf}, we describe the sensor fusion concept with an Extended Kalman Filter (EKF). 
In Section~\ref{sec:results}, we demonstrate the capabilities of the proposed localization system in simulation. These results are discussed in Section~\ref{sec:discussion}, where the work is concluded, summarizing the key aspects and outlining future work.

\section{LiDAR based Odometry}\label{sec:lidar}
Using the point cloud measurements of a LiDAR sensor, movement of the vehicle can be estimated. For this, in \cite{Schichler2025} we used the KISS-ICP (keep it small and simple - iterative closest point) algorithm~\cite{kissicp}. 
For the tunnel environment, we use the GenZ-ICP (Generalized Iterative Closest Point) \cite{genZ} which is developed specifically for degenerative environments such as long corridors. The LiDAR sensor in this demonstration features $1024\times 128$ points with a $45^\circ$ vertical field of view (fov) and a $360^\circ$ horizontal fov.

In the general ICP odometry, each two consecutive frames are matched to estimate the translation between the frames. More precisely, for two consecutive frames with point clouds $\boldsymbol{x_{11}}, \dots,\boldsymbol{x_{kn}}$ and $\boldsymbol{x_{(k+1)1}}, \dots,\boldsymbol{x_{(k+1)n}}$, find the translation $\boldsymbol{t_{k}}$ and rotation $\boldsymbol{R_k}$ that minimize the sum of the squared error:
\begin{equation}\label{eqn:pointmatching}
   \boldsymbol{R_k^*, t_k^*} = \underset{\boldsymbol{R_k, t_k}}{\arg\!\min} \frac{1}{N_p} \sum_{i=1}^{N_p}||\boldsymbol{x_{ki}- R_k\cdot x_{(k+1)i} - t_k}||^2
\end{equation}
Where $\boldsymbol{x_{ki}}$ and $x_{(k+1)i}$ are the point $i$ at time $k$. For finding the points $i$ that are associated to each other, different approaches exist. The GenZ-ICP algorithm relies on the ICP method ~\cite{icp}. 

The Iterative Closest Point (ICP) algorithm relies on the assumption that each point in one point cloud corresponds to its nearest neighbor in the succeeding point cloud. It then iteratively computes a transformation minimizing the point distances. In a planar environment such as corridors or tunnels, it may be beneficial to not minimize the distance between matching points, but the points distances to a plane fitting the measurement. The GenZ-ICP minimizes both the point-to-point and the point-to-plane residuals, with a weighting factor in between that scales with the characteristics of the environment.
% The Iterative Closest Point (ICP) algorithm relies on the assumption that each point in one point cloud corresponds to its nearest neighbor in the succeeding point cloud. It then iteratively computes a transformation minimizing the point distances. However, this approach fails with large motions because the nearest-neighbor assumption does not hold.
% To solve this, we initialize ICP with a constant velocity assumption for predicting the translation between two frames.
% The initial estimate for the translation $(R_k, t_k)$ is
% \begin{subequations}
%     \begin{align}        
%     \boldsymbol{R_k^0} &= \boldsymbol{R_{k-1}}\\
%     \boldsymbol{t_k^0} &= \boldsymbol{R_{k-1}}^T\cdot \boldsymbol{t}_{k-1}
%     \end{align}
% \end{subequations}
% To improve performance, the point cloud is down-sampled with a 3D voxel grid, where only one single point per voxel is kept.

% From the translation estimates $\boldsymbol{R_k}, \boldsymbol{t_k}$, we can compute the velocity $v_k$ and the yaw rate $\omega_k$, using the Riemann Identity $exp(\boldsymbol{R}(\theta))= \theta/(2\cdot\sin\theta)$:%MDPI: Footnote is not permitted in our journal. We moved it to the maintext. Please confirm.
% \begin{subequations}
%     \begin{align}
%         v_k &= \frac{||\boldsymbol{t_{k}}||}{\Delta t}\\
%         \omega_k &= \frac{\log(\boldsymbol{R}_k)}{\Delta t}
%     \end{align}
% \end{subequations}

\section{Thermal Camera-Based Odometry}\label{sec:thermal}
The thermal camera supplements LiDAR by operating reliably in low-light and smoky conditions where other sensors fail. 
The thermal camera used in this demonstration provides 16-bit 2D images. 
Similar to \cite{Schichler2025}, we process thermal images using LDSO \cite{ldso}, a feature-based method employing direct stereo techniques\cite{dso}. LDSO estimates pose and corrects drift via loop closure, improving localization accuracy.
Like LiDAR odometry, matches frames to compute transformations but differs fundamentally from KISS-ICP in methodology. Instead of matching the 3D points, it uses the direct approach, matching the projected 2D points. For downsampling, LDSO uses a gradient-based feature selection method. 

In addition to the transformation, also the depth information needs to be estimated. For the initial guess of the transformation, we use the estimates from the extended Kalman filter (EKF), described in \ref{sec:ekf}. The initial guess for the depth of each pixel is obtained from the depth estimate of the previous frame. %In Figure~\ref{fig:featurepoints}, an exemplary frame with feature points is shown. The color of the feature points corresponds to their estimated depth. 
Jointly estimating transformations and depth introduces drift due to their interdependence. To mitigate this, we employ a key frame approach. Rather than processing every consecutive frame, we estimate the transformation between the current frame and the last key frame.

% \begin{figure}[tb]
  
%     \includegraphics[width=0.9\linewidth, trim={0 0.0cm 0 0.0cm},clip]{fig/featurepoints.png}
%     \caption{Feature points in a thermal camera image. The color of the  points corresponds to their estimated depth.\label{fig:featurepoints}}
% \end{figure} 

%%%%%%%%%%%%%%%%%%%%%%%%%%%%%%%%%%%%%%%%%%
\section{Sensor Fusion for positioning and state estimation}\label{sec:ekf}
We incorporate two sensor systems for the estimation of motion: the LiDAR based odometry and the thermal camera based odometry. To fuse this data, and to estimate the position and orientation of the vehicle, we use an EKF.

The Kalman Filter \cite{kalman} is a well-known approach for the fusion of measurements for state estimation. A very similar form was introduced by \cite{bucykalman}, which has been independently developed at the same time.
The EKF~\cite{ekf0, ekf} is a modification of the Kalman filter fo rnonlinear systems, first used for spacecraft navigation. The EKF has become a widely adopted solution across multiple domains, including mobile robot localization \cite{kalmanlocalizationsurvey}, economic modeling and prediction \cite{stockprediction}, and autonomous system sensor fusion \cite{kalmanrobotsurvey}.

We use the EKF in a loosely coupled framework, implementing a constant acceleration model and performing measurement updates as data from each sensor become available. This is a very common approach for the localization of ground, underwater, or aerial vehicles (e.g., \cite{reviewer}). 
%Building on prior work, such as LVIO-SAM and LVI-SLAM, our approach incorporates thermal imaging and novel adaptations to the LDSO algorithm. These include bilateral filtering to reduce camera noise and integrating EKF-estimated states into keyframe matching to improve depth estimation. The detailed methodology and evaluation presented in this paper highlight the system's capability to achieve accurate and reliable localization under the challenging conditions typical of hazardous environments.

%Simultaneous location and mapping (SLAM) is a key ingredient for autonomous vehicle navigation in unknown environments. Many different SLAM algorithms have been proposed, each involving different levels of complexity and types of sensors. One very simple and yet effective method has been proposed by [cite kiss icp here]. While performing well for relatively short ranges, long-range applications require additional methods to account for drift due to error propagation. The aim is to fuse LiDAR-SLAM data with GNSS data in order to improve the pose estimation of the vehicle. An Extended Kalman Filter with a constant velocity and constant turn rate process model is used for this task.

Similar to the method in \cite{pseudolinearmeas}, we define linear pseudo-measurements to fit the sensor data into the EKF framework.
We define the state vector to be estimated:
\begin{equation}
    \boldsymbol{x}=\begin{bmatrix}x & y & v & \dot v & \psi & \dot\psi & \ddot \psi\end{bmatrix}^T
\end{equation}
where $x,y$ is the global position in the simulation coordinates, $v$ is the velocity, $\dot v$ is the acceleration, $\psi$ is the global yaw angle, $\dot\psi$ is the yaw rate and $\ddot \psi$ is the yaw acceleration. Unlike \cite{Schichler2025}, which employs a constant velocity model, we adopt a constant acceleration model by incorporating $\dot v$ and $\ddot \psi$ into the state vector. This formulation better captures dynamic maneuvers, making it applicable not only to car-like vehicles but also to agile platforms such as legged robots, where motion profiles are more complex and time-varying.

We define the nonlinear dynamic model describing the vehicle's movement by a constant acceleration model, which allows fast changes of velocity and yaw rate.
\begin{equation}
        \boldsymbol{\dot x} = g(\boldsymbol{x}) =  \begin{pmatrix}
        v\cdot\cos\psi &
        v\cdot \sin\psi&
        \dot v&
        0&
        \dot\psi&
        \ddot\psi&
        0
    \end{pmatrix}^T
\end{equation}

For the EKF implementation, discretize in time, using the taylor series expansion.

\begin{subequations}
    \begin{align}
    \begin{split}
        \mathbf{x}_{k+1} &= \mathbf{x}_k + Ts\cdot \left. g(\boldsymbol{x_k})\right|_{\boldsymbol{x}=\boldsymbol{x}_k} + \frac{T_s^2}{2} \cdot\left.\frac{\partial g(\boldsymbol{x})}{\partial t}\right|_{\boldsymbol{x}=\boldsymbol{x}_k} \\&+ \frac{T_s^3}{6}\cdot \left.\frac{\partial^2 g(\boldsymbol{x})}{\partial t^2}\right|_{\boldsymbol{x} = \boldsymbol{x}_k} 
         + \mathbf{w}_{k}
    \end{split}
     \\
    \mathbf{y}_{L,k+1} &= h_L(\mathbf{x}_k) = \begin{bmatrix}
        0 & 0 & 1 & 0 & 0 & 0 & 0\\
        0 & 0 & 0 & 0 & 0& 1 & 0
    \end{bmatrix}\cdot \mathbf{x}_k  + \mathbf{v}_{L,k}\label{eqn:hl}\\
    \mathbf{y}_{T,k+1} &= h_T(\mathbf{x}_k) = \begin{bmatrix}
        0 & 0 & 1 & 0 & 0 & 0 & 0\\
        0 & 0 & 0 & 0 & 0 & 1 & 0
    \end{bmatrix}\cdot \mathbf{x}_k  + \mathbf{v}_{T,k}\label{eqn:ht}
    \end{align}
\end{subequations}
where $\mathbf{y}_{L}$  %MDPI: Please confirm the regular italic/bold format, the formula is italics/bold, and the body text is regular. Please confirm whether a unified format is required. The following highlights are same. 
and $\mathbf{y}_{T}$ are the linear pseudo-measurements. They provide information from the LiDAR and the thermal camera. $\mathbf{w}_k$ is the process noise and $\mathbf{v}_{L,k}$ and $\mathbf{v}_{T,k}$ are the measurement noise for each sensor.

We assume that the process noise $\mathbf{w}_k$ and the measurement noise $\mathbf{v}_{i,k}$ of each sensor $i$ are normally distributed with covariance matrices $\mathbf{R}_k$ and $\mathbf{Q}_{i,k}$, respectively:
\begin{subequations}
    \begin{align}
        \mathbf{w_k} &\sim \mathcal{N}(0, \mathbf{R}_k) \\
        \mathbf{v_{i,k}} &\sim \mathcal{N}(0, \mathbf{Q}_{ik})
    \end{align}
\end{subequations}

The %MDPI: Please confirm whether the unindented format should be retained. The following highlights are same.%Resolved.
 covariance matrices $\mathbf{R}_k$ and $\mathbf{Q}_{i,k}$ quantify the reliability of the dynamic vehicle model and the sensor input, respectively. Model uncertainties (such as, for example, the constant velocity instead of considering the changing velocity) increase $\mathbf{R}_k$, and measurement uncertainties increase $\mathbf{Q}_{i,k}$. Accordingly, tuning these matrices will result in a state estimation tightly following the system dynamics (small $\mathbf{R}_k$) or the sensor data (small $\mathbf{Q}_k$).
% with the nonlinear motion prediction function $g(\mathbf{x_k})$ and the measurement function $h(\mathbf{x_k})$. The measurement function $h$ serves as a map from state space to measurement space. As a prediction model $g$, we use a constant velocity and turn rate (CVTR) model, which is a reasonable approximation for sufficiently small time steps $\Delta t$ between two instances of measurements. 
The EKF procedure as described in \cite{SimonEKF} consists of two steps: the prediction, yielding an a priori state $\mathbf{x^{-}}$, and the correction step, yielding a posterior state $\mathbf{x}^{+}$. 
The prediction step is given as
\begin{subequations}
\begin{align}
    \mathbf{P}_{k+1}^{-} &= \mathbf{G}_k \mathbf{P}_k^{+} \mathbf{G}_k^\top + \mathbf{R}_k \\
    \mathbf{x}_{k+1}^{-} &= g(\mathbf{x}_k^{+})
\end{align}
\end{subequations}
where $\mathbf{G_k}=\partial g(\mathbf{x_k})/\partial \mathbf{x_k}$ is the Jacobian of the system dynamics.
The prediction step is executed until the time step $k$, where new sensor data are received. Then, for this sensor input, a correction step is executed:
\begin{subequations}
\begin{align}
    \mathbf{K}_{k+1} &= \mathbf{P}_{k+1}^{-} \mathbf{H}_{i,k+1}^\top (\mathbf{H}_{i,k+1}\mathbf{P}_{k+1}^{-} \mathbf{H}_{i,k+1}^\top + \mathbf{Q}_{i,k+1})^{-1} \\
    \mathbf{P}_{k+1}^{+} &= (\mathbf{I} - \mathbf{K}_{k+1}\mathbf{H}_{i,k+1})\mathbf{P}_{k+1} \\
    \mathbf{x}_{k+1}^{+} &= \mathbf{x}_{k+1}^{-} + \mathbf{K}_{k+1}(\mathbf{y}_{i,k} - h_i(\mathbf{x}_{k+1}^{-}))
    \end{align}
\end{subequations}
where $\mathbf{H}_{i,k} = \partial h_i(\mathbf{x_k}) / \partial \mathbf{x_k}$ is the Jacobian of the measurement function of sensor $i$. With the applied method of linear pseudo-measurements, $\mathbf{H}_{i,k}$ is equal to the matrices in \eqref{eqn:hl} and \eqref{eqn:ht}.

%%%%%%%%%%%%%%%%%%%%%%%%%%%%%%%%%%%%%%%%%%
\section{Implementation and Test Results}\label{sec:results}

\subsection{Test Setup}
\begin{figure}
    \centering
    \includegraphics[width=0.9\linewidth]{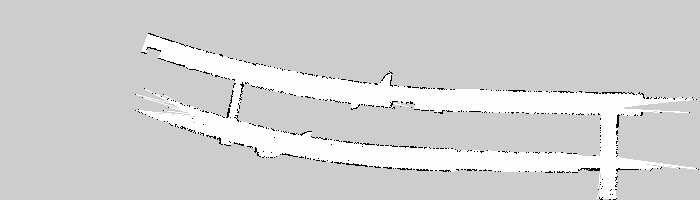}
    \caption{Top--down 2D view of the map used as the simulation environment.}
    \label{fig:map}
\end{figure}

The multi-sensor fusion is simulated in Ignition Gazebo. The map has been generated from a LiDAR pointcloud in the test site Zentrum am Berg \cite{zab}.
The sensors used for the evaluation are models of the u-blox ZED-F9P, Ouster OS2 LiDAR, and the Flir ADK thermal camera. In Figure~\ref{fig:map}, a 2D cutout of this map, which is used for the round trip is shown. 

% \begin{figure}[tb]

%     \includegraphics[width=0.5\linewidth]{fig/map.png}
%     \caption{Top--down view of the driven round trip for the physical data acquirement.}
%     \label{fig:map}
% \end{figure}

For testing the different sensor combinations, we consider the following cases.

%what sensors, where do we drive, sampling frequencies...

%describe ground truth (offline and wiht localization)

\subsection{Ground-Truth}
From our testing ground, an HD map is created in advance. With the Autoware localization stack \cite{locstack}, offline localization is accomplished with high precision. This localization is considered ground-truth data for our evaluation.

% The Groundtruth information is acquired from an offline localization algorithm.
%As the scenario is a commonly used in-house for evaluation, the ground truth is known via the offline localization algorithm from the autoware localization stack \cite{locstack}. A high precision of this ground truth is accomplished by mapping the area previously and fusing a lot of different sensors in the offline localization processing.

\subsection{Test Results}

The GNSS outage is quite common in GNSS-denied environments such as the used tunnel. In such scenarios, the localization task has to be performed with an alternative approach, e.g., by utilizing the LiDAR and thermal camera-based localization solutions.  As both are odometry algorithms, the drift increases over time, especially for the thermal camera odometry, as it has no depth measurement as input.

\begin{figure}
    \centering
    \includegraphics[width=0.9\linewidth]{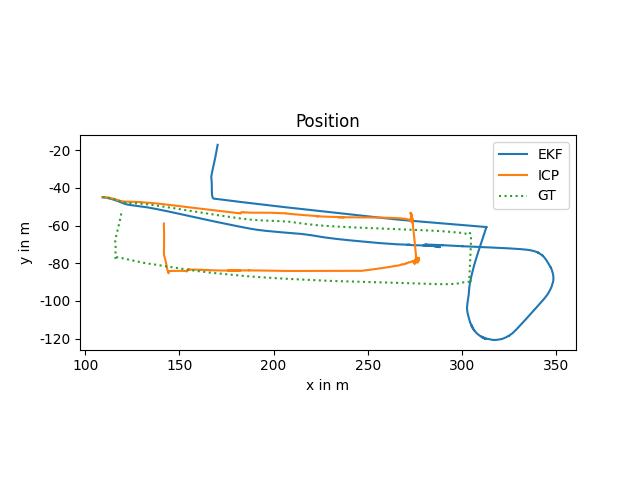}
    \caption{The position estimate of the kalman filter is shown in addition the GenZ-ICP Odometry is shown to highlight the influence of the thermal camera odometry. Lying underneath is the ground truth trajectory shown.}
    \label{fig:ICPTHE}
\end{figure}

\begin{figure}
     \centering
     \includegraphics[width=0.9\linewidth]{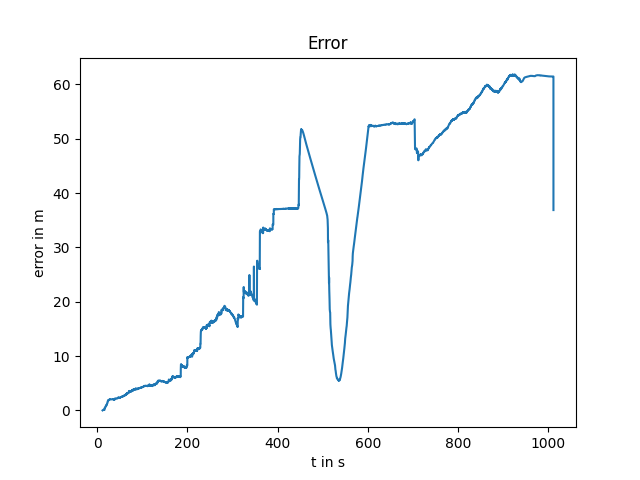}
     \caption{Position error between the ground truth and the estimated trajectory.}
     \label{fig:ICPTHE_err}
 \end{figure}

In Figure~\ref{fig:ICPTHE}, the estimated trajectory utilizing this approach is shown in blue color with the clear drift. The resulting position error over time is shown in Figure~\ref{fig:ICPTHE_err}.

% \begin{figure}[tb]
   
%     \includegraphics[width=0.55\linewidth]{fig/ICPTHE.png}
%     \caption{Localization%MDPI: Please use commas to separate thousands for numbers with five or more digits (not for four digits) in the picture. e.g., "10000" should be "10,000"
%  trajectory where the LiDAR and thermal camera odometry are used. The position estimate of the LiDAR odometry (ICP) is shown in addition to highlighting the influence of the thermal camera odometry.}
%     \label{fig:ICPTHE}
% \end{figure}

\begin{figure}
     \centering
     \includegraphics[width=0.9\linewidth]{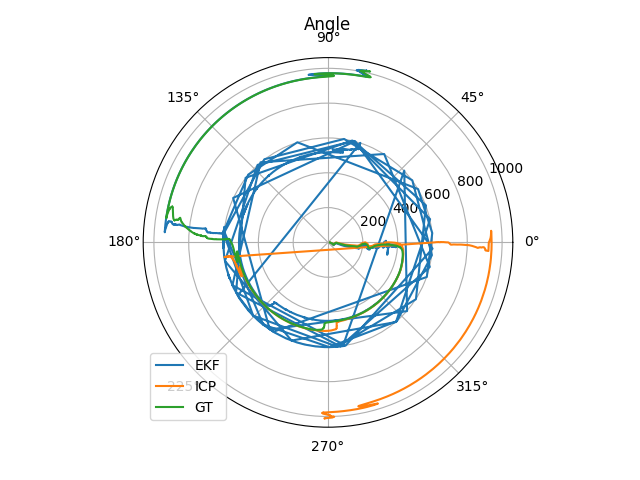}
     \caption{Estimated angle of the extended Kalman filter, and the LiDAR odometry, as well as the ground truth.}
     \label{fig:ICPTHE_phi}
 \end{figure}

When looking at Figure~\ref{fig:ICPTHE_phi}, the reason for the large drift becomes clear, as the angle estimation of the kalman filter after around 10 minutes drifts significantly. Additionally, the thermal camera odometry misestimates the velocity as it has no depth perception.

Although the GenZ-ICP is designed for long corridors and tunnel environments even it has difficulties in this simulation environment as there are next to none features for accurate odometry estimation.
The thermal camera odometry works with a different domain of information allowing it to estimate poses even when there a none 3D features for the lidar odometry to work with.

% \begin{figure}[tb]

%     \includegraphics[width=0.55\linewidth]{fig/ICPTHE_err.png}
%     \caption{Position error between the ground truth and the estimated trajectory.}
%     \label{fig:ICPTHE_err}
% \end{figure}

% \begin{figure}[tb]

%     \includegraphics[width=0.55\linewidth]{fig/ICPTHE_phi.png}
%     \caption{Estimated %MDPI: Please confirm whether the overlapped numbers effects reading, if so, please revise.
%  angle of the thermal camera odometry, and the LiDAR odometry, as well as the estimation of the extended Kalman filter.}
%     \label{fig:ICPTHE_phi}
% \end{figure}


\newpage

%%%%%%%%%%%%%%%%%%%%%%%%%%%%%%%%%%%%%%%%%%
\section{Discussion and Future Work}\label{sec:discussion}

In this paper, we propose fusing LiDAR and thermal camera data using an extended Kalman filter to ensure robust localization in hazardous or GNSS-denied settings. In doing so, we show complementary strengths, where LiDAR provides 3D odometry, thermal imaging mitigates issues like fog, smoke or 3D feature less environments. 

The multi-sensor fusion of LiDAR, and thermal camera odometry opens up localization for a wide range of scenarios due to the complementary advantages and disadvantages of the individual sensors. The presented method of sensor fusion with an extended Kalman filter comprises a simple and modular implementation with easy extendability and intuitive parameter tuning. The sensor fusion between GNSS and LiDAR odometry is common practice for autonomous vehicles but is not resilient to many hazardous environmental impacts like airborne particles (e.g., fog and smoke), tunnels, or skyscraper environments. Using additional odometry information, especially provided by sensors like a thermal camera with complementary strengths and weaknesses, extends the fields of applications. However, odometry with only one image at a time is a challenge, and the accuracy of localization is lower than other localization methods in many cases.

Despite the addition of numerous modifications to the original monocular visual odometry algorithm, its accuracy enhancements fall short of facilitating precise localization in challenging environments. This is partly a result of the environment's scale being unobservable, which causes velocity and rotation estimates to be inaccurate. 

As both algorithms are not capable of providing global position information such a GNSS system would do, the Lidar Odometry can be further improved by extending the algorithm with a Loop-Closure feature which enables a higher accuracy in the pose estimation over longer distances. 

Additionally, the internal Lidar Odometry optimization algorithm simply uses a constant velocity prediction model, which can be extended via feedback from the kalman filter itself or with an additional IMU. As the optimization algorithm is also only dependent on the sequence of pointclouds information over the prediction variance could lead to more accurate pose estimation. With such changes to the lidar odometry and maybe also to the LDSO would shift the algorithm into the tightly coupled domain.

To fully utilize the kalman filter expension to the acceleration domain an IMU can also be used as an adiitional sensor in order to improve the pose estimation. Here the advantage of the kalman filter being loosely coupled enables for fast sensor integration.

\end{document}